%% file: main.tex
\documentclass{article}
\usepackage{acra}
\usepackage{amsfonts}  

\usepackage{graphicx}

\usepackage{siunitx}  
\usepackage{booktabs}  

\title{Towards Self-Attention Based Visual Navigation in the Real World}
\author{Jaime Ruiz-Serra\textsuperscript{1,*}, 
        Jack White\textsuperscript{1}, 
        Stephen Petrie\textsuperscript{1}, \\          
        {\bf Tatiana Kameneva\textsuperscript{1,2}}, \and 
        {\bf Chris McCarthy\textsuperscript{1}} \\
        \textsuperscript{1}Swinburne University of Technology, Australia \\
        \textsuperscript{2}Melbourne University, Australia \\ 
        \textsuperscript{*}jruizserra@swin.edu.au}

\begin{document}

\maketitle

\begin{abstract}
\input{0-Abstract}
\end{abstract}

\input{1-Introduction}
\input{2-LitReview}
\input{3-Methods}
\input{4-ResultsDiscussion}
\input{5-Conclusions}

\bibliography{library}
\bibliographystyle{named}

\end{document}

%% file: 0-Abstract.tex
Vision guided navigation requires processing complex visual information to inform task-orientated decisions. 
Applications include autonomous robots, self-driving cars, and assistive vision for humans. 
A key element is the extraction and selection of relevant features in pixel space upon which to base action choices, for which Machine Learning techniques are well suited. 
However, Deep Reinforcement Learning agents trained in simulation often exhibit unsatisfactory results when deployed in the real-world due to perceptual differences known as the \emph{reality gap}. 
An approach that is yet to be explored to bridge this gap is self-attention. 
In this paper we (1) perform a systematic exploration of the hyperparameter space for self-attention based navigation of 3D environments and qualitatively appraise behaviour observed from different hyperparameter sets, including their ability to generalise; (2) present strategies to improve the agents' generalisation abilities and navigation behaviour; and (3) show how models trained in simulation are capable of processing real world images meaningfully in real time. 
To our knowledge, this is the first demonstration of a self-attention based agent successfully trained in navigating a 3D action space, using less than 4000 parameters.

%% file: 1-Introduction.tex
\section{Introduction}\label{introduction}

\begin{figure}[ht]
  \centering
  \includegraphics[width=\linewidth]{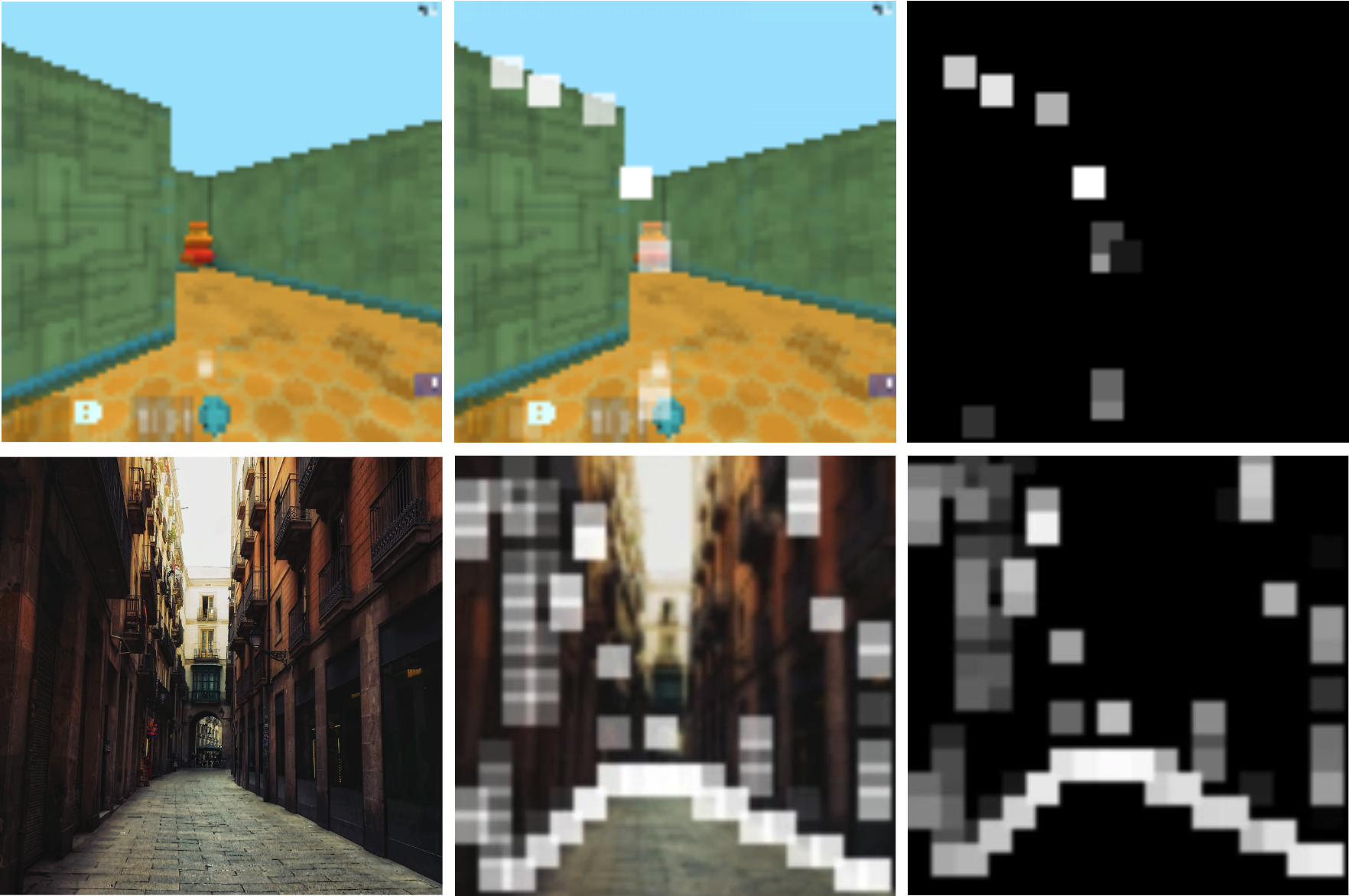}
  \caption{Trained agents are capable of selecting important features of the environment in both simulation and the real world, in real time. \textit{Left Column}: unprocessed frames. \textit{Center Column}: the $K$ most important patches overlaid on frames. \textit{Right Column}: the $K$ most important patches in isolation. \textit{Top Row}: frame from simulation training environment, $K=10$. \textit{Bottom Row}: real-world image, $K=80$.}
  \label{fig:main-figure}
\end{figure}

Vision-based navigation is a challenging task that requires perceiving and rapidly processing highly complex information to determine actions.
In the context of vision-guided robotics, visual navigation requires extracting features of the environment that directly inform behaviour choices, and/or or serve as direct inputs to control in order to reach an intended goal or destination.  More broadly, such vision processing may also serve to support human activity in tele-operation contexts, as well as in assistive technologies for human mobility. 

The problem of navigation can be framed as a function that maps sensor-acquired input information at a given time step (e.g.~camera-captured pixels), to an action choice that both progresses towards a pre-determined goal, while also satisfying constraints such as avoiding collision, and minimising distance travelled. 
This framing entails the reduction of a high-dimensional space into a single vector of actions, which Reinforcement Learning (RL) algorithms are particularly well suited to do. 
These algorithms are capable of learning complex behaviours over many iterations of loss or fitness evaluation and parameter adjustments, requiring a repeatable, controlled environment in which to train. 
Training in simulation offers multiple advantages over training in the real world in terms of speed, repeatability, safety, and ability to modify the environment. 
Once controllers are trained or evolved, they need to be appraised for performance---and eventually deployed---in the real world.

The difference in complexity between simulation and real-world
environments, known in the literature as the \emph{reality gap}, is a major challenge that controllers trained in simulation face when deployed in the real world \cite{Tobin2017}. 
The most common approaches to narrow the gap are to increase the fidelity of the simulation environment to bring it ``closer'' to the real world \cite{Bousmalis2018,Stein2018}, and randomising visual aspects of the simulation with the intention to have the real world seem like just another variation of the simulation \cite{Tobin2017,Tremblay2018}. 
The additional computational steps required by these methods increases the time required to train agents \cite{Zhang2019}.

Biological vision employs saccades to gather information in sequence, subsequently contrasted with (and used to update) our conceptual models of the world \cite{Khetarpal2018}. 
This gives rise to \emph{inattentional blindness}, whereby humans focus only on important parts of the visual scene, largely ignoring the rest \cite{Simons1999}. Ignoring task-irrelevant information reduces the complexity of the inputs that need to be processed. 
From this angle, instead of learning to adapt to differences by using larger models, we can have agents learn to focus on the most important parts of the input, which are likely to be similar across a range of different environments (both simulation and real-world).

\emph{Self-attention}, a means to compute sequence representations based on internal relationships popularised by~\cite{Vaswani2017}, can be used to enforce an information bottleneck to remove task-irrelevant parts from agent inputs, mimicking inattentional blindness.
Additionally, self-attention is more parameter- and compute-efficient \cite{Tang2020}, making it more suitable for hardware-constrained applications.

In this paper we show the potential for self-attention as a means to bridge the reality gap for real-world navigation. 
Specifically, we assess the feasibility of a simulation-trained self-attention model to generate simple representations of real-world states that are relevant to the task of navigation. 
We thus adapt the agent from \cite{Tang2020} to the context of vision- and depth-based navigation of 3D spaces. 
The agent, as in the original implementation, is trained by means of \emph{neuroevolution}, a class of evolutionary algorithms used to iteratively improve neural networks based on a fitness score. 
This is possible due to the low parameter count facilitated by self-attention.

In summary, we (1) perform a systematic exploration of the hyperparameter space for self-attention based navigation of 3D environments and qualitatively appraise behaviour observed from different hyperparameter sets, including their ability to generalise; (2) present strategies to improve the agents' generalisation abilities and navigation behaviour; and (3) show how models trained in simulation are capable of processing real world images meaningfully in real time. 
To the best of our knowledge, this is the first self-attention based agent successfully trained in navigating a 3D action space.

%% file: 2-LitReview.tex
\section{Background}\label{background}

\emph{Reinforcement Learning} (RL) is a class of ML algorithms wherein agents learn to perform actions based on the observed state of the environment they exist in and the expected future rewards of available actions. 
To perform vision-based RL, algorithms must be capable of processing images as environment observations (i.e. high-dimensional inputs).
To this end, \textit{Deep RL} (DRL), originally presented in \cite{Mnih2013}, incorporates deep learning algorithms into the RL paradigm.
The most popular DRL algorithms for vision-based navigation are model-free: value-based (Deep Q-Learning \cite{Mnih2013} and variants) or policy-based (TRPO \cite{Schulman2015} and variants). 
A limitation of these algorithms is their requiring fully-differentiable components in order to be trained using gradient descent.

The complexity of DRL models requires thousands of iterations to train them, thus training them in real world systems is prohibitively slow. 
Instead, DRL models are usually trained in simulation environments and later deployed in the real world. 
This offers several advantages: training episodes can be considerably sped up beyond real time, allowing for many training iterations at a marginal extra cost; episodes are undertaken in a controlled, repeatable environment; agents are not at risk of damaging the environment or themselves; environments can be modified for different tests \cite{Jakobi1998}. 

On the other hand, a major disadvantage of simulation environments is their lack of fidelity in modelling the real world. 
This discrepancy in observations, known as the \emph{reality gap}, poses a challenge for simulation-trained models acting in the real world \cite{Zhang2019}. 
Increasing the fidelity of the simulation environment is the intuitive approach \cite{Bousmalis2018}. 
Other approaches include domain randomisation \cite{Tobin2017,Peng2018}; abstraction of sensory inputs and actions \cite{Scheper2017}; combining simulation with off-policy data \cite{Bharadhwaj2019}; and Hebbian plasticity and online adaptation \cite{Qiu2020}. 
Of relevance to our study is the approach of separating perception and decision-making, by downsampling and encoding the input images to a lower-dimensional latent space (perception) for use by a downstream controller (decision-making), e.g. \cite{Lobos-Tsunekawa2018,Ma2019,Gordon2019,Tang2020}.


Feature extraction is an essential element of vision processing.
In machine learning-driven work, the most tried and tested visual feature extraction method are Convolutional Neural Networks, as they encode rich and complex visual representations.
Despite their history as the standard vision processing primitive, their receptive fields do not scale well enough to model spatially long-range dependencies in the inputs \cite{Ramachandran2019a}. 
Further, the number of parameters scales quadratically with input size, quickly increasing the computational cost \cite{Tang2020}.

The RL agent proposed by \cite{Tang2020} that we further develop here incorporates a bottleneck to mimic inattentional blindness. 
This bottleneck is achieved by means of SA as an importance voting mechanism on image patches and subsequently selecting only the most important regions of the image for action decision-making. 
By ignoring irrelevant parts of the input, the hypothesis is that the agents should be more robust to variations in the environments visual attributes and thus generalise better, albeit at a cost in performance \cite{Khetarpal2018}.
This makes self-attention a promising method for narrowing the reality gap.
Along similar lines, \cite{Khetarpal2018} use a saliency filter to attenuate parts of the images deemed less salient. 
\cite{Lehnert2019}, use a retina-inspired module before a convolutional neural network to navigate 3D simulation environments. 
SA represents a highly parameter-efficient solution that is suitable in low-computational-resource settings, as well as evolutionary strategy training \cite{Tang2020}.

\emph{Neuroevolution} algorithms apply evolutionary optimisation to neural networks.
It has been proven to be ``effective in training a feature extractor'' to be used in conjunction with other ML techniques \cite{Verbancsics2013}.
The main advantage of evolutionary processes is that they are able to train architectures containing non-differentiable components \cite{Tang2020}. 
Additionally, as a Monte Carlo method they are well suited to distributed computing, thus being more scalable.
This is an advantage over other existing DRL training methods.

%% file: 3-Methods.tex
\section{Methodology}\label{methodology}

Our approach is centered around the self-attention agent proposed by
\cite{Tang2020}, described in more detail in this section. Their work,
in a similar fashion to \cite{Mott2019}, is limited to 2D simulation
environments. To obtain results that are relevant to our use case of
real world navigation, we adapted the method for 3D navigation, trained
and tested the agent in a 3D environment, and subsequently evaluated its
performance on real-world images.

\subsection{Simulation Environment}\label{simulation-environment}

We train the agent in \emph{DeepMind Lab} (DML) \cite{Beattie2016},
which offers 3D environments with differences in appearance and task
goals. The agent's task is to navigate through a maze to reach a goal,
collecting apples along the way. Rewards are considerably sparser in DML
as compared with the environments used by \cite{Tang2020} and
\cite{Mott2019}, forcing the agent to learn temporally long-range
consequences of its actions. Agents are trained in one of
\texttt{NavMazeStatic01} or \texttt{NavMazeRandomGoal03} environments,
and tested in others as discussed in Section \ref{training-and-testing}.

\emph{Observation space:} The environment generates an observation at
each timestep, in the form of a tensor
$\mathbf{O}_t \in \mathbb{R}^{64\times 64\times C}$, with $C=3$ for
RGB and $4$ for RGB-D. We train RGB and RGB-D agents, as well as
agents with depth-only observations ($C=1$) and agents with grayscale
+ depth observations ($C=2$); and explore the differences in their
performance.

\emph{Action space:} For this study, we use an action space
dimensionality of three (\emph{look left/right}, \emph{strafe
left/right}, \emph{move forward/backward}), thus reducing the agent's
parameter space and thereby facilitating the search for an optimal set
of parameters in training. Our three-dimensional action space is larger
than that of \cite{Tang2020}. This, together with higher reward sparsity
and observation variability, represents a considerably more challenging
task for the agent to solve, bringing results closer to the ultimate
goal of real-world navigation.

\emph{Episode length and frame skip:} The number of observations in a
training episode is determined by the episode length, the frame rate
(fps), and the frame skip rate. The agent ``sees'' every fourth
observation (frame skip), holding actions for that long. We use a 60 sec
training episode length at 60 fps for direct comparison with other work
(e.g. \cite{White2019,Khetarpal2018,Lehnert2019}),
resulting in $(60\times 60) / 4 = 900$ observations/episode.

\emph{Rewards:} Agents receive a reward of 10 points each time they
reach a goal, and 1 point for each apple collected along the way. To
address the sparsity of rewards in the environment, and with the
intention to correct some of the undesirable behaviours exhibited by
agents, we conducted experiments incorporating an exploration reward
($r_{expl}$) whereby the agent gets rewarded each time it visits a
location in the training environment for the first time, as determined
by a grid; as well as penalties (i.e.~negative rewards) each time the
agent collides with a wall ($p_{col}$), as in \cite{Brunner2018}.
Baseline agents are configured with $r_{expl} = 0.1$ and
$p_{col} = 0$.

\subsection{Self-attention Agent}\label{self-attention-agent}

The agent, introduced by \cite{Tang2020}, partitions the input images
into patches and ranks the patches by their perceived importance by
means of self-attention. All action control decisions by the agent are
made based solely on the \emph{location} of the $K$ most important
patches within the image.

\emph{Preprocessing and partitioning the input:} The observations
produced by the task environment at each time step are $(64, 64, C)$
pixel tensors. These values are normalised and partitioned into patches,
each of shape $(M, M, C)$, using a sliding window method. Patches are
employed in order to reduce the number of computations performed by the
agent with each observation. The number of patches from an input image
is given by $n = (\lfloor\frac{L-M}{S}+1\rfloor)^2$, with input image
size $L$, stride $S$, and patch size $M$. They are subsequently
flattened into $d_{in}$-dimensional vectors and combined into a single
matrix $\mathbf{X} \in \mathbb{R}^{n \times d_{in}}$. Agents that
utilise the depth channel in observations convert depth values into
disparity values (inverse of depth) prior to partitioning into patches.

\emph{Ranking by salience:} The self-attention mechanism, in combination
with \emph{softmax}---explained in detail in \cite{Tang2020}--- acts as
a voting system for the importance of patches. At its core are two
matrices of learnable parameters
$W_k, W_q \in \mathbf{R}^{d_{in} \times d}$, where $d$ is the
dimension of the latent space and has an integer value of small
magnitude. The voting yields an importance vector assigning an
importance score to each patch. Only the $K$ most important patches
are used in later parts of the network.

\emph{Feature retrieval:} Having selected the $K$ most important
patches by means of self-attention, features at the patch locations can
be retrieved to serve as inputs to the controller. We follow the
original agent implementation \cite{Tang2020} in using the
\emph{location} of the patches on the input image only.

\emph{Controller:} The agent's controller is a single-layer, 16-cell
LSTM that combines the self-attention output with its hidden state. Its
output is passed through a fully connected layer, yielding a vector of
normalised action values for the actions defined. The normalised action
values are interpolated to the action ranges and rounded to the nearest
integer prior to their application to the environment.

\subsection{Training and Testing}\label{training-and-testing}

The agents were trained by means of neuroevolution with the CMA-ES
algorithm \cite{Hansen2016}, as in \cite{Tang2020}. Agents were evolved
over 1200 generations, with a population of 64 members and 8 training
episodes per member per iteration. To increase the scalability and efficiency of training, we completely decoupled the CMA population from the training task queue. 
Task requests, including population member identifier and agent parameters for the given population member are placed in a queue and undertaken by compute workers on a FIFO basis. 
This makes the training more flexible and suitable for distributed computing. 

Upon training, we selected the agents with best performance and tested them in each of the DML \texttt{NavMaze} environments, consisting of mazes of different sizes, color schemes and textures, where the goal may be always in the same location (\texttt{Static}), or randomly placed (\texttt{RandomGoal}). 
The visual differences amongst the environments make them good candidates to test the agents' ability to generalise.
The agents that showed better generalisation abilities were tested for processing real-world scenes using images and video. 
We show simulation score distributions and processed real-world images for select agents in Section \ref{results-and-discussion}.

%% file: 4-ResultsDiscussion.tex
\section{Results and Discussion}\label{results-and-discussion}



All experiments involve training and testing agents with the
architecture from \cite{Tang2020}, adapted to 3D navigation. In this
section, we use the term \emph{agent} to refer to an instance of the
self-attention agent configured by a given set of hyperparameters. The
hyperparameters within the scope of this paper are summarised in Table
\ref{tab:agent-configs}, agent \texttt{C3} being our baseline agent. We
present quantitative analysis of the learning curves of the agents and
their performance in different environments, and a qualitative
assessment in pixel space of the agents' learnt patch selection and
resultant behaviours.

We found that self-attention agents are capable of navigating 3D spaces
effectively, with very efficient training. Agents reached scores above
\(60\) in \texttt{NavMazeStatic01} with under \(0.1\mathrm{E}9\)
training steps, and were able to navigate environments with different
visual qualities to the training environment. As shown in Figure
\ref{fig:main-figure}, agents learn to select image patches that are
relevant to the task of navigation in both simulation environments and
real-world images.

\subsection{Simulation Environments}\label{simulation-environments}

\begin{figure*}[!ht]
  \centering
  \includegraphics[width=\linewidth]{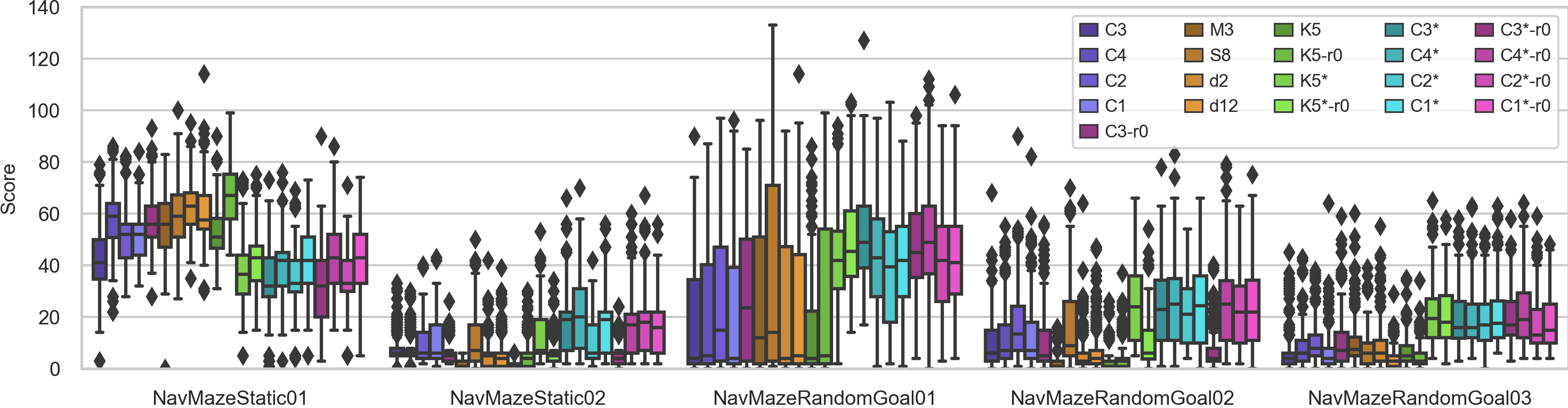}
  \caption{Agent score distributions over 200 trials in each selected DML environment. Key hyperparameters for each of the agents are listed in Table \ref{tab:agent-configs}. Agents marked with ``*'' were trained in \texttt{NavMazeRandomGoal03}. All other agents were trained in \texttt{NavMazeStatic01}.}
  \label{fig:agents-boxplot}
\end{figure*}

We tested the agents in both the training environment and other
environments with similar tasks but different visual and structural
qualities. We observed their scores, patch selection, and behaviour.
Overall, agents are able to translate the representations learnt in the
training environment to other environments. Results are shown in Figure
\ref{fig:agents-boxplot}.

\setlength{\textfloatsep}{0.1cm}

\begin{table}[t]
\label{tab:agent-configs}
\centering
\begin{tabular}{ l r r r r r r r }
    \toprule
    Agent & $C$ & $M$ & $S$ & $d$ & $K$ & $r_{expl}$ & $p_{col}$ \\
    \midrule
    C3 (baseline) & 3 & 7 & 4 & 4 & 10 & 0.1 & 0 \\
    C4 & \textbf{4} & 7 & 4 & 4 & 10 & 0.1 & 0 \\
    C2 & \textbf{2} & 7 & 4 & 4 & 10 & 0.1 & 0 \\
    C1 & \textbf{1} & 7 & 4 & 4 & 10 & 0.1 & 0 \\
    C3-r0 & 3 & 7 & 4 & 4 & 10 & \textbf{0} & 0 \\
    C3-p0.02 & 3 & 7 & 4 & 4 & 10 & 0.1 & \textbf{-0.02} \\
    M3 & 3 & \textbf{3} & 4 & 4 & 10 & 0.1 & 0 \\
    S8 & 3 & 7 & \textbf{8} & 4 & 10 & 0.1 & 0 \\
    d2 & 3 & 7 & 4 & \textbf{2} & 10 & 0.1 & 0 \\
    d12 & 3 & 7 & 4 & \textbf{12} & 10 & 0.1 & 0 \\
    K5 & 3 & 7 & 4 & 4 & \textbf{5} & 0.1 & 0 \\
    K5-r0 & 3 & 7 & 4 & 4 & \textbf{5} & \textbf{0} & 0 \\
    \bottomrule
\end{tabular}
\caption{
    Agent hyperparameters selected for comparison. 
    $C$: input channels, 
    $M$: patch size, 
    $S$: stride, 
    $d$: latent space dimensionality, 
    $K$: number of patches selected, 
    $r_{expl}$: exploration reward, 
    $p_{col}$: collision penalty. 
    More details in Sections \ref{simulation-environment} and \ref{self-attention-agent}. 
    Agent names are based on changed hyperparameters, highlighted in \textbf{bold}.
    }
\end{table}

\setlength{\textfloatsep}{0.7cm}

\emph{Depth perception:} Results demonstrate that adding depth to data
input enabled the agent to focus on structural characteristics of the
environment over more specific features of the DML environment such as
its highly contrasting walls/floor/sky planes. This results in more
``human-like'' patch selection of apples, floor (as opposed to
floor/wall boundaries), and vertical corner edges. Agents are less
distracted by the patterns on the walls and the score panels on the
bottom part of the screen. Depth-only agents (\(C=1\)) proved to be
competitive in generalising to other environments, both from simpler to
more complex environments and vice versa. When comparing agents with
both depth and RGB/grayscale perception (\(C=4\) and \(C=2\)), 4-channel
agents trained in difficult environments (i.e.
\texttt{NavMazeRandomGoal03} which is larger, more complex and has
sparser, more randomised rewards than \texttt{NavMazeStatic01})
generalise better to other environments, but 2-channel agents trained in
easier environments generalise better to difficult environments. This may be
because by removing color information, the observation distributions are
closer between environments.

\emph{Patch overlap:} The patch size \(M\) and stride \(S\) have a
direct impact in the number of trainable parameters, and therefore the
complexity of training. Smaller patches and larger strides result in
less overlap between patches as well as less parameters to be learnt.
Agents with no overlap achieved scores as high as any others, and showed
better generalisation to other environments. This may be due to the fact
that no overlap prevents the selected patches to be in clusters over a
single region of the image, forcing them to be distributed in a
grid-like manner over the image.

\emph{Latent space:} The size of the self-attention matrices \(d\)
determines the complexity of single patch representations used for
importance voting. A higher-dimensional latent space does not result in
higher scores or better generalisation.

\emph{Patch count:} Smaller \(K\) obtains higher scores, but it appears
to cause the agent to overfit to the training environment, as
generalisability is lower. This aligns with the fact that the LSTM controller
receives a smaller input.

\begin{figure}
  \centering
  \includegraphics[width=\linewidth]{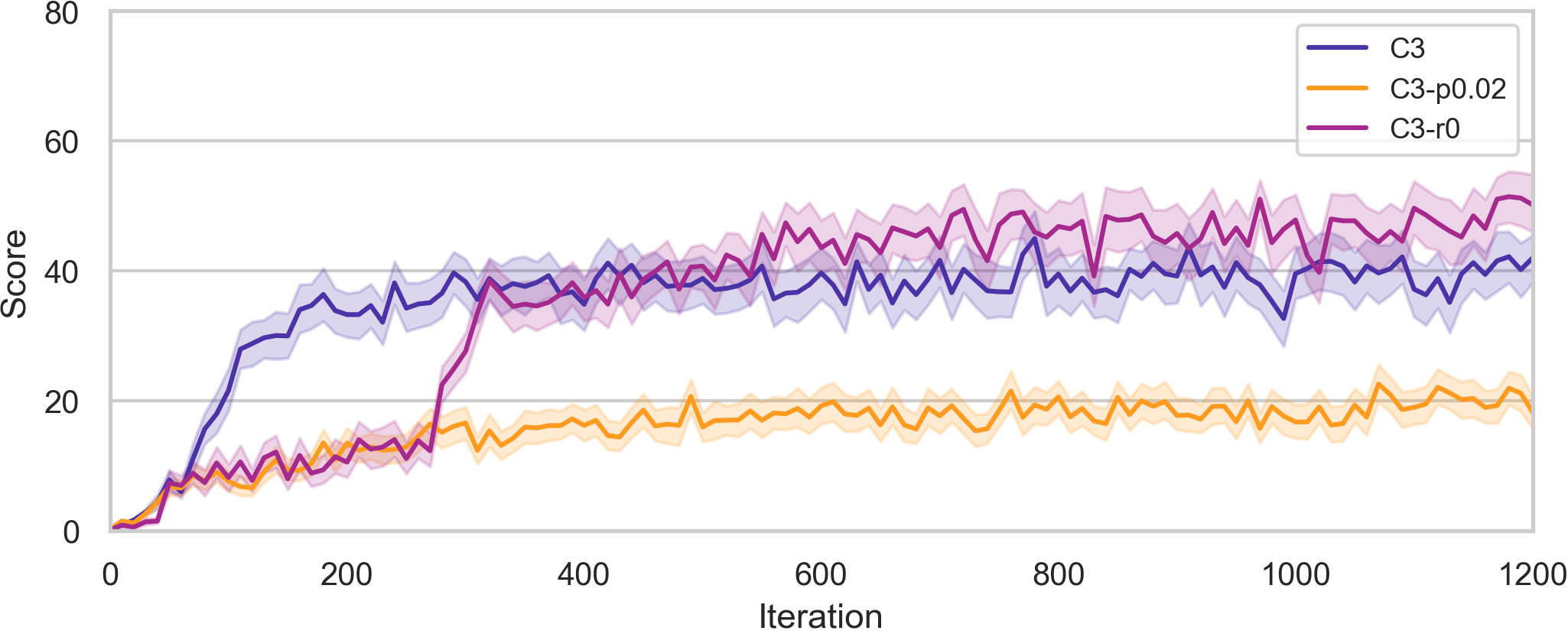}
  \caption{Training progress in \texttt{NavMazeStatic01} from agents with different rewards. Exploration rewards help learn faster but result in lower scores in the long run (\texttt{C3}). Collision penalties address behaviour but result in lower scores (\texttt{C3-p0.02}).}
  \label{fig:learning-curve-plot}
\end{figure}

\emph{Exploration rewards} \(r_{expl}\) seem to help agents learn faster
than those without, with total reward plateauing at high levels about 2
times faster during training on \texttt{NavMazeStatic01} (score plots in
Figure \ref{fig:learning-curve-plot}), and about 4 times faster during
training on \texttt{NavMazeRandomGoal03}. This suggests that curiosity
rewards are particularly helpful in larger, more complex environments
with sparser rewards. On the other hand, agents with exploration reward
obtained lower scores overall, suggesting that exploration rewards may
be beneficial early on in the training process, but detrimental in the
long run. Exploration reward does not affect generalisability, with the 
exception of RGB agents trained on \texttt{NavMazeRandomGoal03}. The 
addition of exploration reward enhances these agents' generalisation 
ability in larger levels: when tested on \texttt{NavMazeStatic02} 
and \texttt{NavMazeRandomGoal02}, \texttt{K5*} and \texttt{C3*} outperform 
their \texttt{r0} counterparts, as shown in Figure \ref{fig:agents-boxplot}. 

Agents with smaller \(K\) trained quicker without exploration reward
than agents with exploration reward, suggesting limitations of the
controller. Further work on self-attention agents may focus on
addressing the network's LSTM controller, with an emphasis on its
inputs.

\emph{Collision penalties} \(p_{col}\) hinder learning speed, overall
scores, and generalisation. Agents trained from scratch with collision
penalties successfully learn to avoid collisions with walls, but this
results in undesired ``stuck in place'' behaviour. Extending agent
training with collision penalties resulted in marginal improvements in
training environment performance.

Additional rewards or penalties such as the ones described above may be
helpful in curriculum learning settings, with exploration reward
speeding up the initial part of the training, and collision penalties
addressing agent limitations later on in the process.

\emph{Training in more complex environments}, in our case
\texttt{NavMazeRandomGoal03}, resulted in agents achieving better
generalisation (marked ``*'' in Figure \ref{fig:agents-boxplot} legend).
Our assertion is that the randomisation of the goal location, in
addition to the increased structural variation of the environment,
resulted in policies that relied on general environmental features more
heavily, and thus overfitted less to the environment they were trained
on. This supports the idea of using self-attention in combination with
other methods such as domain randomisation or more realistic
environments to bridge the reality gap.

\subsection{Real-world Images}\label{real-world-images}

\begin{figure*}[t!]
  \centering
  \includegraphics[width=\textwidth]{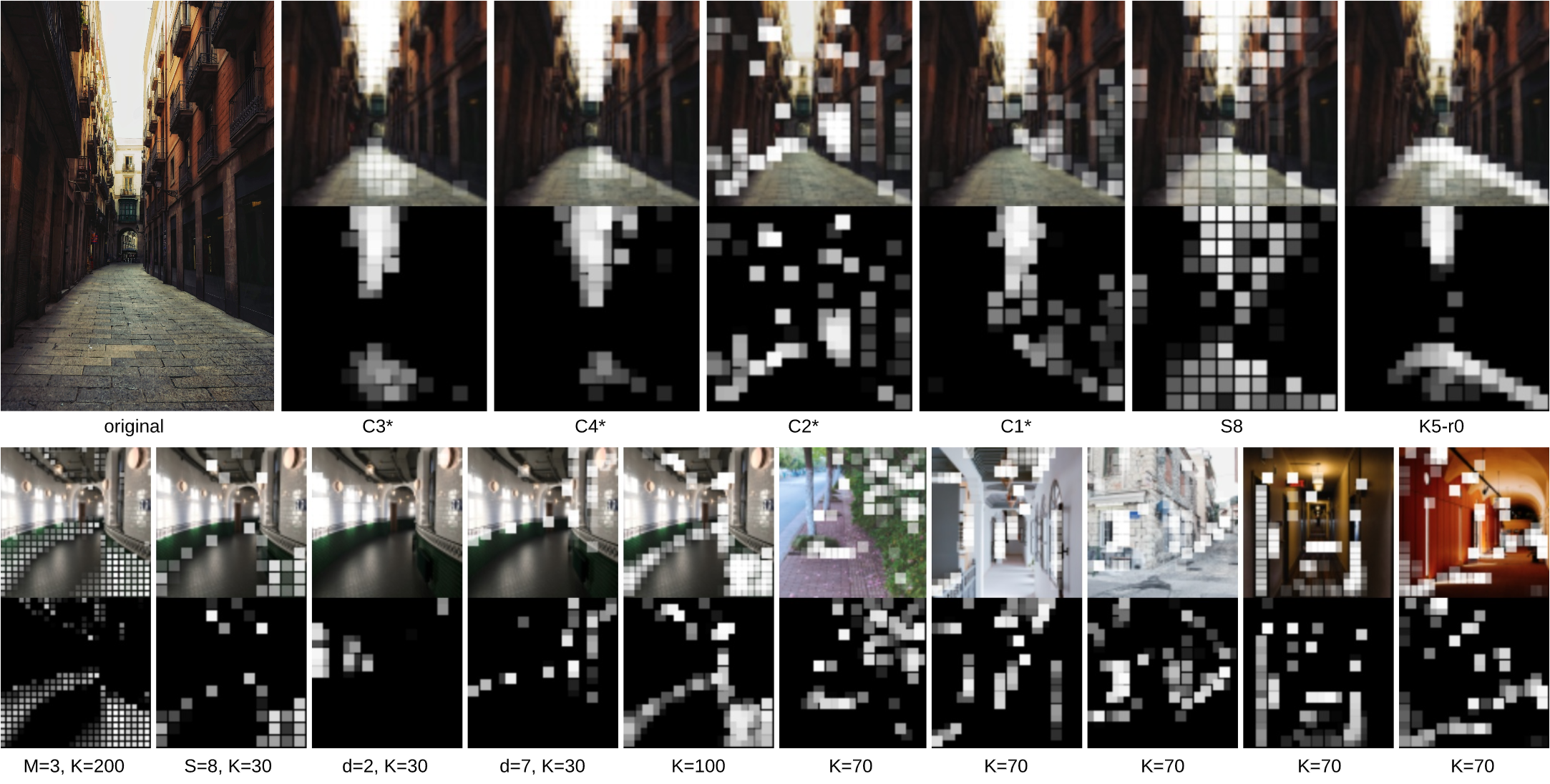}
  \caption{Real-world images processed by the self-attention network. Agents trained in simulation are able to select useful features from real-world images with different color schemes and contrast levels. Patch intensities map to their relative importance. \textit{Top:} Different agents processing the same image for comparison, with agent names. \textit{Bottom:} Images processed by agent \texttt{C2*}, with hyperparameters changed post training as described in the legend.}
  \label{fig:real-world-compared}
\end{figure*}

We found that processing real images using our best performing agents
effectively highlights task-relevant features in the real world. Sample
results are shown in Figure \ref{fig:real-world-compared}, where it can
be seen that, despite the image being of low contrast, and of a
completely different color scheme to that of the training environment,
the model is able to rank and select the most important regions of the
image. These can be further processed with other feature retrieval
methods including depth sampling or convolutions to increase the amount
of information they convey to downstream parts of the system. In
addition to this, the temporal dimension in non-static inputs
(i.e.~video) increases interpretability considerably. 

Worth noting is the ability to change hyperparameters (e.g. \(K\)) of already-trained
models at image processing time, in real time. This can be performed adaptively by later
parts of the system, increasing the expressivity of the self-attention
component. Examples of this are shown in Figure \ref{fig:real-world-compared} (bottom two rows).

%% file: 5-Conclusions.tex
\section{Conclusions and Future Work}\label{conclusions-and-future-work}

In this paper, we demonstrate that self-attention agents trained in
simulation are capable of selecting relevant visual features and using
them to navigate 3D maze environments, being, to our knowledge, the
first to do so. The agents, comprising less than 4000 learnable
parameters, learned sensible navigation behaviours with less than
\(0.1\mathrm{E}9\) training environment steps. Furthermore, we show that
the agents' learnt patch selection can be used to identify useful
features in real-world scenes, proving to be a promising approach to aid
in bridging the reality gap.

Self-attention agents may be used in combination with other approaches
from the reality gap literature---such as more realistic simulations and
domain randomisation---and as a saliency or attention layer to be added
to more complex agent architectures. Both suggestions are to be explored
in future work, along with further agent enhancements. Specifically,
depth information and/or convolutional layers may be used in the feature
retrieval part of the network, providing richer information to the LSTM
controller.